\documentclass{article}

\usepackage{PRIMEarxiv}

\usepackage[utf8]{inputenc} 
\usepackage[T1]{fontenc}    
\usepackage{hyperref}       
\usepackage{url}            
\usepackage{booktabs}       
\usepackage{amsfonts}       
\usepackage{nicefrac}       
\usepackage{microtype}      
\usepackage{lipsum}
\usepackage{fancyhdr}       
\usepackage{graphicx}       
\graphicspath{{media/}}     

\usepackage{amsmath,amssymb,amsfonts}
\usepackage{algorithmic}
\usepackage{graphicx}

\usepackage{algorithm}
\usepackage{xcolor}
\usepackage[english]{babel}

\usepackage{comment}

\usepackage{tkz-euclide}
\usepackage{pst-plot}

\usepackage{caption}
\usepackage{subcaption}
\usepackage{multirow}

\usepackage{tikz}
\usepackage{pgfplots}
\pgfplotsset{compat = newest}

\usetikzlibrary{spy}

\newcommand{\R}{\mathbb{R}} 
\newcommand{\dd}{\mathrm{d}} 

\title{Retrospective Motion Correction in Gradient Echo MRI by Explicit Motion Estimation Using Deep CNNs}

\author{
  Mathias S. Feinler, Bernadette N. Hahn \\
  Department of Mathematics \\
  University of Stuttgart \\
  Germany\\
  \texttt{\{mathias.feinler, bernadette.hahn\}@imng.uni-stuttgart.de} \\
  }

\begin{document}
\maketitle

\begin{abstract}
	Magnetic Resonance Imaging allows high resolution data acquisition with the downside of motion sensitivity due to relatively long acquisition times. Even during the acquisition of a single 2D slice, motion can severely corrupt the image. Retrospective motion correction strategies do not interfere during acquisition time but operate on the motion affected data. Known methods suited to this scenario are compressed sensing (CS), generative adversarial networks (GANs), and motion estimation. In this paper we propose a strategy to correct for motion artifacts using Deep Convolutional Neuronal Networks (Deep CNNs) in a reliable and verifiable manner by explicit motion estimation. The sensitivity encoding (SENSE) redundancy that multiple receiver coils provide, has in the past been used for acceleration, noise reduction and rigid motion compensation. We show that using Deep CNNs the concepts of rigid motion compensation can be generalized to more complex motion fields. Using a simulated synthetic data set, our proposed supervised network is evaluated on motion corrupted MRIs of abdomen and head. We compare our results with rigid motion compensation and GANs.

\end{abstract}


\section{Introduction}
Suitable algorithms for motion correction are highly relevant for medical imaging applications such as Magnetic Resonance Imaging (MRI). In MRI we have the possibility to choose the acquisition scheme. Conventional spin echo sequences require long preparation times between echos (at each echo, limited data can be measured). With multishot sequences we can acquire an echo train (a limited number of echos) but between these echo trains the aforementioned preparation time is still necessary. With Gradient echo (GRE) sequences a constant data acquisition is feasible as a steady state is formed. The resulting acquisition time still requires that patients do not move for several seconds or minutes.

During data acquisition involuntary motion can occur due to non-cooperative patients or induced by respiratory and cardiac motion. Prospective motion compensation strategies usually require temporal supersampling, regeneration \cite{bydder2002} or acquisition during resting periods \cite{herborn2003}, navigators \cite{kober2011} or tracking \cite{maclaren2013, zaitsev2015} and gating. These strategies are known to be time consuming and prone to errors. Since most procedures are designed for a static setup, the most elegant way of motion compensation is a retrospective one. 

The acquisition schemes determine if or to which extend retrospective motion compensation is feasible. For conventional spin echoes, two contiguous data acquisitions are potentially severely inconsistent due to motion during long preparation times. This renders motion estimation almost impossible in general. For multishot setups, each echo train is acquired in a short period of time. This allows to conclude that the data of one shot is intrinsically almost consistent, but might be severely inconsistent with the data of other echo trains. The intuitive model for multishot setups is therefore that motion during an echo train is negligible. Hence all data of one echo train is assigned to one deformation field. We can transfer this idea to GRE sequences by artificially separating the data, bearing in mind that the continuous data acquisition allows us to balance the temporal resolution retrospectively. 

Strategies for retrospective motion compensation can be divided into multiple categories including acceleration together with compressed sensing (CS), generative adversarial networks (GANs) and explicit motion estimation. 

Faster acquisitions are less sensitive to motion. Sensitivity encoding (SENSE) allows subsampling without compromising the image quality \cite{pruessmann1999}. Further subsampling can be compensated by CS \cite{vranic2018}. A wide range of classical reconstruction algorithms have been proposed e.g. based on total variations (TV) \cite{cruz2016}  and wavelets \cite{lustig2007}. Recently, Machine Learning techniques have produced qualitatively excellent reconstructions in this scenario \cite{adler2017, adler2018, li2019, hammernik2018, schwab2019, hauptmann2020}. 
The reconstruction is in this case dependent on the training data set. In case of strong undersampling, limited angles or similar scenarios, the reconstruction might lead to an erroneous diagnosis while the visual appearance seems flawless.

The temptation of longer acquisitions is that these provide more data and hence more information. However, these data are corrupted by motion, and therefore a static reconstruction comprises motion artifacts. The idea of GANs is to produce an artifact free image from the static reconstruction \cite{usman2020, armanious2018b, armanious2018a, lunz2018}. 
Motion can cause features to appear or vanish within the static reconstruction. Additionally the reference configuration is not unique anymore. These issues are tackled by a discriminator network that is learned to distinguish between corrupted and artifact free images. The issue with this approach is that data consistency is not guaranteed anymore. For vanilla GANs this problem is known as mode collapse \cite{goodfellow2016}. Even for conditional GANs this issue can still appear. Hence the problem of appearing or vanishing features is not solved satisfactory.

The concept of implicit motion estimation allows to state higher assumptions like periodicity or low-dimensionality in time on the motion without explicitly computing deformation fields. These assumptions can be used to improve the reconstruction of motion affected data by UNFOLD \cite{madore1999}, k-t-blast, k-t-SENSE or k-t-focuss \cite{tsao2003, jung2007}. Since deformations are not calculated, instead the temporal variation of a series of reconstructions is used to penalize high temporal frequencies. This concept is highly dependent on the appearance of the reconstructions. Hence, in cases where reconstructions contain complex details or deformations are non-periodic, this approach is limited.

The computationally most expensive yet most promising approach in retrospective motion compensation is the explicit use of deformation fields. These allow to compute fully data consistent reconstructions. However, these deformation fields are rarely available and need to be estimated. Therefore different approaches have been proposed. For rigid motion one can minimize the residual value by a newton procedure w.r.t the deformation parameters \cite{cordero2016}. Even moderate sized problems require many iterations and the convergence is increasingly uncertain for strong deformations and disadvantageous sampling trajectories. Therefore, this idea has been extended and accelerated with low resolution scout scans (SAMER \cite{polak2021}), neuronal networks (NAMER \cite{haskell2019}) and model reduction (TAMER \cite{haskell2018}). Clearly this approach is limited to simply parameterized deformation fields only, since the Hessian needs to be computed.

One approach to reach more complex deformation fields is a variational model for joint motion estimation and image reconstruction \cite{burger2018, chen2019}. Such descriptions are particularly hard to solve and require a certain finesse to identify hyperparameters and the right constraints. These penalize the temporal and spatial non-smoothness of deformation fields and noise within the reconstructed image. Usually for compliant deformation fields the optical flow \cite{horn1981} constraint is used. Even for a single 2D slice solving the optimization requires rather hours than minutes.

The use of Neuronal Networks has led to significantly decreased computation times \cite{haskell2019}. In that sense also the problem of image registration has been revolutionized by the use of Deep CNNs. In the literature, there are many methods \cite{oliveira2014, haskins2020} trained in supervised \cite{sokooti2017} and unsupervised \cite{balakrishnan2019} fashion, for uni- and multi-modal setups \cite{hu2018}, with one-shot approaches like FlowNetS \cite{fischer2015} and iterative \cite{ilg2016} strategies. Supervised methods usually use optical flow or synthetic deformations as a ground truth while unsupervised methods penalize the difference of the morphed images and the non-smoothness of the deformation fields. All these approaches require two fully resolved images to be compared to each other. In the setup of motion estimation within the acquisition of a single 2D slice, only partial k-space information is available at each point in time to estimate the deformation state. Further, if artifact affected images are fed into unsupervised methods, deformation fields are produced which translate the artifacts as well. One idea is the use of local all-pass filters to contruct a suitable network for non-rigid registration (LAPNet) \cite{kuestner2021}. This method outperforms FlowNetS when the same sub-sampling operator is used for both moving and reference image. In general, this condition on the subsampling operator is not satisfied. 
Thus the direct relation in k-space cannot be used anymore. Hence, the ideas of the literature have to be adapted to be applicable in this scenario.

In this paper we propose a new learned procedure to deduce deformation fields in GRE MRI. In a first step, we separate the data temporally. Each block can then be used to calculate individual undersampled reconstructions. By comparison of these reconstructions a first estimate to the global deformation fields can be inferred. To this end, a suitable network architecture is proposed. The main contribution of this paper is then a proposed subsequent learned bi-level gradient descent algorithm to compensate for remaining deformation field inaccuracies. The combined framework allows to retrospectively estimate and compensate for motion artifacts in MRI. To train these networks we synthetically generate ground truth deformation fields and simulate the according motion corrupted data acquisition. The framework can hence be trained in a supervised fashion. 

Our networks generalize the idea of Cordero et al. in \cite{cordero2016} to virtually arbitrary deformation fields by transferring the concepts of Adler et al. in \cite{adler2017} of iterative deep neuronal networks for solving ill-posed problems to motion estimation.

The remainder of this paper is organized as follows. Section II introduces the theoretical basis including the forward model, k-space trajectory and the key ideas of our motion estimation approach. In Section III the structure of the proposed networks is presented along with details on network training and the discrete setup. The results on simulated brain and abdominal MRIs are shown in Section IV and a conclusion is drawn in Section V.

\section{Theoretical Basis}

\subsection{Forward Imaging Model} 
The overall goal is to recover a reference image $s^\mathrm{ref}$ and a sequence of deformation fields $U^\mathrm{ref}$ from measured MRI data $y$. Therefor, a suitable forward operator is required, to model the imaging process. We use the motion model for multishot MRI of Batchelor et al. \cite{batchelor2006}
\begin{equation}
	\mathcal{A}(U,s) = \sum_{i=1}^{N^{\mathrm{exc}}} \sum_{c=1}^{N^{\mathrm{coils}}} A_{i} \mathcal{F}[S_c U_{i}[s]].  \label{eq:model} \nonumber
\end{equation}
Here, $N^\mathrm{exc}$ stands for the number of excitations, $U = \{U_{i}\}_{i=1}^{N^\mathrm{exc}} \in \mathcal{D}_U$ is the (a priori unknown) sequence of deformation fields contained in the space of admissible deformations $\mathcal{D}_U$, $s \in \mathcal{D}_s$ is the searched-for image at reference configuration contained in the space of admissible ground truth images $\mathcal{D}_s$, $\mathcal{F}$ is the Fourier transform and $A_{i}$ is the characteristic function for the k-space portion extracted at excitation time $t_i$. The $N^\mathrm{coils}$ coil sensitivity maps $\{S_c\}_{c=1}^{N^\mathrm{coils}}$ are assumed to be known by some precomputation procedure \cite{allison2012} and are time-invariant.

The space of admissible deformations $\mathcal{D}_U$ is a generic space that is implicitly defined in context with each application. For brain MRIs $\mathcal{D}_U$ might be limited to rigid deformations, whilst for abdominal free breathing MRIs much more complex deformations that contain the totality of possible deformations caused by human breathing define $\mathcal{D}_U$. Such abstract definitions are in most cases impossible to tackle exactly. Therefore, $\mathcal{D}_U$ can be extended to an easier parameterizable superspace such that elements can be drawn. The space of admissible ground truth images $\mathcal{D}_s$ is to be understood in the same sense.

In general, measured data $y = \mathcal{A}(U^\mathrm{ref},s^\mathrm{ref}) + \mathcal{N}$ are only available with noise. We model this measurement noise $\mathcal{N}$ as additive normal distributed noise. We can split $y$ into contributions per excitation and coil sensitivity map by $y = \{y_i\}_{i=1}^{N^\mathrm{exc}} = \{\{y_i^c\}_{c=1}^{N^\mathrm{coils}}\}_{i=1}^{N^\mathrm{exc}}$.

$U_{i}[s]$ resembles the deformed state of the ground truth image at excitation time $t_i$. Subsection \ref{subsec:subsection_deformation_model} will further elaborate how these deformations are modelled. Within this model it is intrinsically assumed that the deformation state does not change over the period of one excitation but only in between. 

We further define the field of view $\Omega^\mathrm{FOV} $ as a two-dimensional rectangle.

\subsection{Deformation and Image Model}\label{subsec:subsection_deformation_model}

Actually measured nonlinear deformation fields are hardly available ground truths for most applications. Therefore we need to model rich enough synthetic deformation fields such that all naturally occurring features of the respective application are captured. We assume that free form deformations \cite{rueckert1999} are general enough to capture most effects. 
Therefor a regular grid $\{x_{l,m}\}_{l,m=1}^{N^\mathrm{M}}$ with $x_{l,m} \in \R^2$ is introduced. At each time $t_i$ and for each node $x_{l,m}$ a support vector $v_{l,m} \in \Omega^\mathrm{FOV}$ is defined by choosing an affine deformation $v_{l,m} = \Gamma_{l,m}(t_i)x_{l,m} + b_{l,m}(t_i)$ where $b_{l,m}(t_i) \in \R^2$ and $\Gamma_{l,m}(t_i) \in \R^{2 \times 2}$. The deformation $U_{i}$ can now be defined as the spline interpolation of the support vectors onto the field of view. Therefore, $U_{i}$ contains absolute positional references, i.e. the particle at position $x$ at time $t_i$ is at reference configuration located at position $U_{i}(x)$.

Note, that the spline interpolation of first order provides continuous piecewise affine deformations. If for all $l,m$ $\Gamma_{l,m}(t_i) = \Gamma(t_i)$ and $b_{l,m}(t_i) = b(t_i)$, the free form deformation reduces to a global affine deformation. If further $|\mathrm{det}(\Gamma(t_i))| = 1$, this model only allows for signal intensity preserving inplane shifts, rotations and shearing.

Free form deformations are, however, lacking the possibility of modeling discontinuous deformations. These can appear extensively in thoracic or abdominal cavity due to respiratory motion and the induced movement on organs. 

To model these effects properly, an arbitrary strictly convex set $C$ with $\mathcal{C}^1$ boundary is placed inside the field of view. The motions inside and outside of $C$ are modelled independently as free form deformation $U^{\mathrm{in}}$ and $U^{\mathrm{out}}$, respectively. The inner deformation is constrained such that no mass can enter or leave that convex set. This is established by first removing the contributions normal to the boundary of $C$. 

For all $(r_1,r_2)^\top \in \partial C$ let $n_{r_1,r_2}$ denote the normal vector of the boundary of $C$. Further let $U^{\mathrm{in},\perp}_{r_1,r_2} := n^\top_{r_1,r_2} (U^{\mathrm{in}}_{r_1,r_2} - (r_1,r_2)^\top)$ be the normal contribution. For all $(r_1,r_2) \in C$ we define
\begin{align}
	\bar{U}^{\mathrm{in}}_{r_1,r_2} := U^{\mathrm{in}}_{r_1,r_2} &-  U^{\mathrm{in},\perp}_{r_1^t(r_2),r_2}(r_1-r_1^b(r_2))/(r_1^t(r_2)-r_1^b(r_2)) \nonumber \\ 
	&-  U^{\mathrm{in},\perp}_{r_1^b(r_2),r_2}(r_1-r_1^t(r_2))/(r_1^t(y)-r_1^b(r_2)) \nonumber
\end{align}
where $r_1^t(r_2) := \max_{u\in \R}\{(u,r_2)^\top \in \partial C\} $ and $r_1^b(r_2) = \min_{u\in\R}\{(u,r_2)^\top \in \partial C\}$. By strict convexity these points are unique. Furthermore we set $\bar{U}^{\mathrm{in}}_{r_1,r_2} = (r_1,r_2)^\top$ for all $ (r_1,r_1)^\top \in \Omega^\mathrm{FOV}\backslash C$. Further, we have to assure that $\mathrm{supp}( \bar{U}^{\mathrm{in}}[s_{|C}]) \subseteq C$, where $s_{|C} := s \chi_{C}$ with the characteristic function $\chi_{C}$ of the set $C$. The full deformation hence reads $U[\cdot] = U^\mathrm{out}[\bar{U}^{\mathrm{in}}[\cdot]]$. 

The explicit definition of the deformations $U^\mathrm{in}$ and $U^\mathrm{out}$ can for example be established by defining $\Gamma_{l,m}$ and $b_{l,m}$ smooth in space and time. We further assume that $U_{1}=I$ i.e. we define the deformation state at the first excitation as the reference configuration. 
With this approach fairly general motion fields can be synthetically generated.

Now, we can draw from this distribution of deformation fields to simulate motion corrupted MRIs and finally train the networks. Please note that in contrast to the approach of Cordero et al. \cite{cordero2016} we do not aim to estimate the explicit parameters used to generate this deformation.

It remains to find a suitable model for $s$. We therefor use artifact free images of the respective application. We assume that this finite data set renders $\mathcal{D}_s$ well enough to capture all characteristics for the application onto unseen data.

\subsection{Sampling Trajectories}
The choice of the sampling trajectory determines e.g. the characteristics of the introduced motion artifacts, the possibility to estimate motion, reliability etc. Therefore, the sampling trajectory has to be balanced carefully to achieve a setup where retrospective motion correction is feasible.

It has been found, that radial sampling is relatively insensitive to motion whereas conventional cartesian sampling can produce strong ghosting \cite{forbes2001}. 
From the literature it is well known that if the spokes are evenly spaced, the performance of undersampled CG-SENSE reconstructions is superior to random or golden angle distributions \cite{chan2011}.
Based on these findings we suggest radial parallel acquisition for the remainder of this paper. 
The temporal dependency of the acquisition is of high significance for motion compensation. Hence, also the chronological order in which the k-space is sampled is highly important. 
We suggest a greedy-approach with respect to the k-space coverage, leading to van der Corput sampling \cite{chan2011}. All ideas nonetheless can be generalized to a larger class of sampling trajectories.

\subsection{Motion Estimation}

Motion estimation requires some redundancy in the measurements. In our case we want to make use of parallel imaging by using multiple receiver coils. The sensitivities of these coil arrays need to be computed. As mentioned above, we assume, that the sensitivity maps do not change during motion corrupted data acquisition and can be measured and computed by a preprocessing step. Additionally these sensitivity maps are assumed to be available at least over the set $\{r \in \Omega^{\mathrm{FOV}} \,\, | U_i(s^\mathrm{ref})(r) > 0, \, i=1,\ldots,N^\mathrm{exc} \}$. This requires a procedure which extrapolates sensitivity maps robustly outside the calibrated region \cite{allison2012, cordero2016}. 

The locality of the sensitivity maps lead to the possibility to assess motion corruption by the residuum
\begin{equation}
	R(U,s) = y - \mathcal{A}(U,s) \label{eq:residuum}
\end{equation}
where for the static forward model the deformations are replaced by the identity operator. If motion occurs, the residuum is in general unequal to zero if the forward operator is equipped with an incorrect deformation field. Therefore the residuum can be used to estimate and validate motion. 

Since by radial acquisition the center of k-space is sampled at each excitation equally well, additionally to the sensitivity maps, this overlap can help to extract motion. From the literature it is known that radial acquisition allows sub-Nyquist sampling rates whereas still almost exact reconstructions can be expected \cite{chan2011}. If on the other hand too few spokes are sampled, a sparse reconstruction scheme needs to be applied. 

Since the center of k-space is sampled at each excitation, the characteristics of the static reconstruction $s_i$ using just $y_i = \{y_i^c\}_{i=1}^{N^\text{coils}}$ remains similar for each $i$. 

The most direct way to estimate motion is therefore a comparison of the static reconstructions $s_i$ of each excitation. 

In that sense we aim to find only estimates $U^{\text{est}}$ by registration of static CG-SENSE reconstructions using $y_i$ which shall then be improved by a subsequent correction step described in the following.

\subsection{Correction of Motion Estimate}\label{subsec:motion_correction}
For the reconstruction step, we assume that we are now equipped with data $\{y_i\}_{i=1}^{N^\mathrm{exc}}$ and deformation estimates $\{U_{i}^{\text{est}}\}_{i=1}^{N^\mathrm{exc}}$ from the registration of individual undersampled CG-SENSE reconstructions. 

Since the data are assumed to be only available with some additive normal distributed noise, the motion estimates can only be expected to be further degraded. If inaccuracies in $U_{i}^\mathrm{est}$ are ignored, the best reconstruction can be expected by minimizing
\begin{equation}
	\min_s \| \mathcal{A}(U^\mathrm{est},s) - y \|_2^2 + \mathcal{R}(s)  \nonumber
\end{equation}
where the explicit choice for the regularization term $\mathcal{R}$ is independent of motion. Hence, the class of static regularized reconstruction algorithms from the literature can now be applied to the linear forward operator $\tilde{\mathcal{A}}(s) = \mathcal{A}(U^\mathrm{est},s)$.

To compute precise reconstructions, we have to take into account that $U^\mathrm{est}$ is obtained by a previous motion estimation step. Thus, we have to account for inaccuracies, i.e. the exact motion field is rather given by $U^\mathrm{est} + \Delta U$ with unknown $\Delta U$. This leads to solving the full optimization problem
\begin{equation}
	\min_{s, \Delta U} \| \mathcal{A}(U^\mathrm{est} + \Delta U,s) - y \|_2^2 + \mathcal{R}(s,U^\mathrm{est} + \Delta U). \label{eq:min_with_Uguess}  \nonumber
\end{equation}
The huge benefit of introducing $U^\mathrm{est}$ is that, presuming some level of accuracy of $U^\mathrm{est}$, a more accurate deformation exists \textit{locally}.

Thus, local optimization techniques like gradient descent converge to the true optimum, if certain regularity assumptions on $s$ and $U$ are stated and a certain accuracy of $U^\mathrm{est}$ is assumed. Empirically, for rigid transformations, Cordero achieved almost sure convergence for small motion and small number of excitations \cite{cordero2016}. However, the results showed very slow convergence-rates for large motion and sometimes a rather sudden drop of error after several dozen iterations. A preliminary guess on deformations would hence significantly reduce runtime and allow almost sure global convergence.

For general motion we can introduce the parameter field $p_{i} = \{p_i^x, p_i^y\}$ parameterizing the motion field $U_{i}$. The derivatives w.r.t $(p_{i})_{j,k}$ of the to be minimized data fidelity term $J(U) := \| \mathcal{A}(U,s) - y \|_2^2$ are given by
\begin{equation}
	\frac{\dd J(U)}{\dd (p_{i})_{j,k} } = \sum_{c=1}^{N^\text{coils}}2\Re( w_{i,c}^* w_{i,c,(j,k)})  \nonumber
\end{equation}
where $\Re(\cdot)$ returns the real part of a complex number, $(\cdot)^*$ denotes the adjoint and
\begin{align}
	w_{i,c}^* &= (A_{i} \mathcal{F}[S_cU_{i}(s)] - y_i^c)^*  \nonumber \\
	w_{i,c,(j,k)} &= A_{i} \mathcal{F}[S_c \frac{\dd U_{i}}{\dd (p_{i})_{j,k} }(s)].  \nonumber
\end{align}

For arbitrarily parameterized deformations, the derivative $\frac{\dd U_{i}}{\dd (p_{i})_{j,k} }(s)$ is in general nonzero everywhere. This is indeed the case for rigid transformations. For brevity we will drop the index $i$ in the following and perform the steps using $p^x$ for illustrative purposes. We assume the parameters are chosen to have local influence by using bilinear interpolations. The crucial property of these locally parameterized deformations is that the derivative
\begin{align}
	\frac{\dd U}{\dd p^x_{j,k} }(s) &= \Bar{U}(\frac{\dd s}{\dd x})(j,k) e_{j,k}, \label{eq:dU/dpx}
\end{align}
where $\Bar{U}$ is parameterized by $\lfloor p^x_{j,k} \rfloor$ and $p^y_{j,k}$, is nonzero only at index $(j,k)$. This leads to the fast implementable gradient

\begin{equation}
	\frac{\dd J(U)}{\dd (p^x_{i}) } = \sum_{c=1}^{N^\text{coils}} 2\Re\left( \frac{\dd U}{\dd p^x_i}(s) \circ S_c^*\mathcal{F}^*[A_{i}^* w_{i,c}]\right) \label{eq:gradients} 
\end{equation}
where $\circ$ denotes the Hadamard product.

For gradient descent schemes one problem is always the choice of the stepwidth. One way to prevent choosing a stepwidth and increasing the order of convergence is using Newton's method by introducing the Hessian matrix. Unfortunately the Hessian matrix is in our case only blockdiagonal w.r.t. $i$, but each block is a full $2N^2 \times 2N^2$ matrix. Hence the inversion of such matrices is time and memory consuming. Still we need to correct for some effects to achieve an almost universal choice of the stepwidth. We therefore employ the diagonal elements of the Hessian only. These elements are given by 
\begin{equation}
	\frac{\dd^2 J(U)}{\dd ((p^x_i)_{j,k})^2 } = \sum_{c=1}^{N^\text{coils}} 2\Re\left( \frac{\dd U}{\dd (p^x_i)_{j,k}}(s) S_c^* K_i S_c \frac{\dd U}{\dd (p^x_i)_{j,k}}(s) \right) \label{eq:Hessian_diagonal}  \nonumber
\end{equation}
with 
\begin{equation}
	K_i = \mathcal{F}^*A_{i}^* A_{i} \mathcal{F}.  \nonumber
\end{equation}
The operator $K_i$ is a symmetric yet full matrix. Since the expression $\frac{\dd U}{\dd (p^x_i)_{j,k}}(s)$ leads to a vector with only one entry unequal to zero, only the diagonal entries of $K$ are relevant. Due to the Fourier shift theorem 
all diagonal elements of $K_i$ are equal. This value can be precomputed as long as $A_{i}$ remains unchanged. 

For radial acquisitions with a symmetrically distributed number of spokes, the operator $K_i$ behaves like the convolution with a blurring kernel. The true inverse can be found by applying the Lucy–Richardson deconvolution \cite{richardson1972} with the known downside of noise amplification . In that sense taking just the inverse of the diagonal elements the result can be seen as a smoothed version of the true inverse of the Hessian. We expect a better noise resistance and much less computation time. Since the derivatives of equation \eqref{eq:dU/dpx} can become zero, also the diagonal elements of the Hessian are not bounded from zero. To prevent any numerical problems, we finally add the unit matrix times the twentieth of the maximum value of all diagonal elements
\begin{equation}
	\Bar{H}_i^x = \frac{\dd^2 J(U)}{\dd ((p^x_i)_{j,k})^2 } + 0.05\left\| \frac{\dd^2 J(U)}{\dd ((p^x_i))^2 } \right\|_{\mathrm{max}}. \label{eq:final_Hessian}
\end{equation}

Even if an analytic intertwined Newton procedure similar to Corderos' can minimize the data fidelity error very well, which also works for our deformation parametrization, the result is on the other hand not favorable. The reason is, that admissible deformations are not necessarily given in terms of the Newton step $U^{(k+1)} = U^{(k)} + (\Delta_U J(U^{(k)}))^{-1} \nabla_U J(U^{(k)}) ) \not \in \mathcal{D}_U$. Hence, conventional gradient descent or Newton steps would lead to artifacts in $U$ and $s$.

Well known solutions to this problems are proximal gradient descent methods and as a special case projected gradient descent methods. For convex problems convergence guarantees and fast convergence rates can be proven, but usually require explicit modelling of regularization terms on $U$ or the explicit definition of $\mathcal{D}_U$, respectively. On the one hand, similar to Cordero, one can expect almost sure convergence, in proximity to the true solution. On the other hand for complex deformations, these tasks are challenging and in addition computationally expensive. 
Instead, we rely on neural networks leading to learned iterative algorithms with proven effectivity \cite{adler2017,arridge2019}.

\subsection{Final Motion-Compensated Reconstruction}

For the computation of a final motion compensated reconstruction the CG-SENSE plus motion procedure can be applied. 

The known relation that rotations of the image correspond to rotations in k-space can lead to undersampling artifacts even if the exact deformation fields are found. Therefore, depending on the application, a sparse reconstruction can be necessary. Using total variations compressed sensing plus motion, we can compensate remaining noise and undersampling artifacts simultaneously. 
In case of negligible undersampling artifacts, also a residual CNN denoiser \cite{zhang2017} can be used in favor of time efficiency.

\section{Methodology} 

\subsection{First Motion Estimation}
The setup for estimating the transformations $U^\text{est}$ directly arises from theory by comparing the CG-SENSE reconstructions $s_i$ to the reference state $s_1$. Since the center of k-space is sampled at each excitation, the basic appearance of all $s_i$ is similar. Since on the other hand different spokes are sampled, the details and artifacts from the violation of the Nyquist criterion lead to different images even if no motion appears. Hence we have a weak multi-modal registration setup and rely on Deep CNNs. 

\begin{figure}
	\centering
	\includegraphics[width=0.3\textwidth]{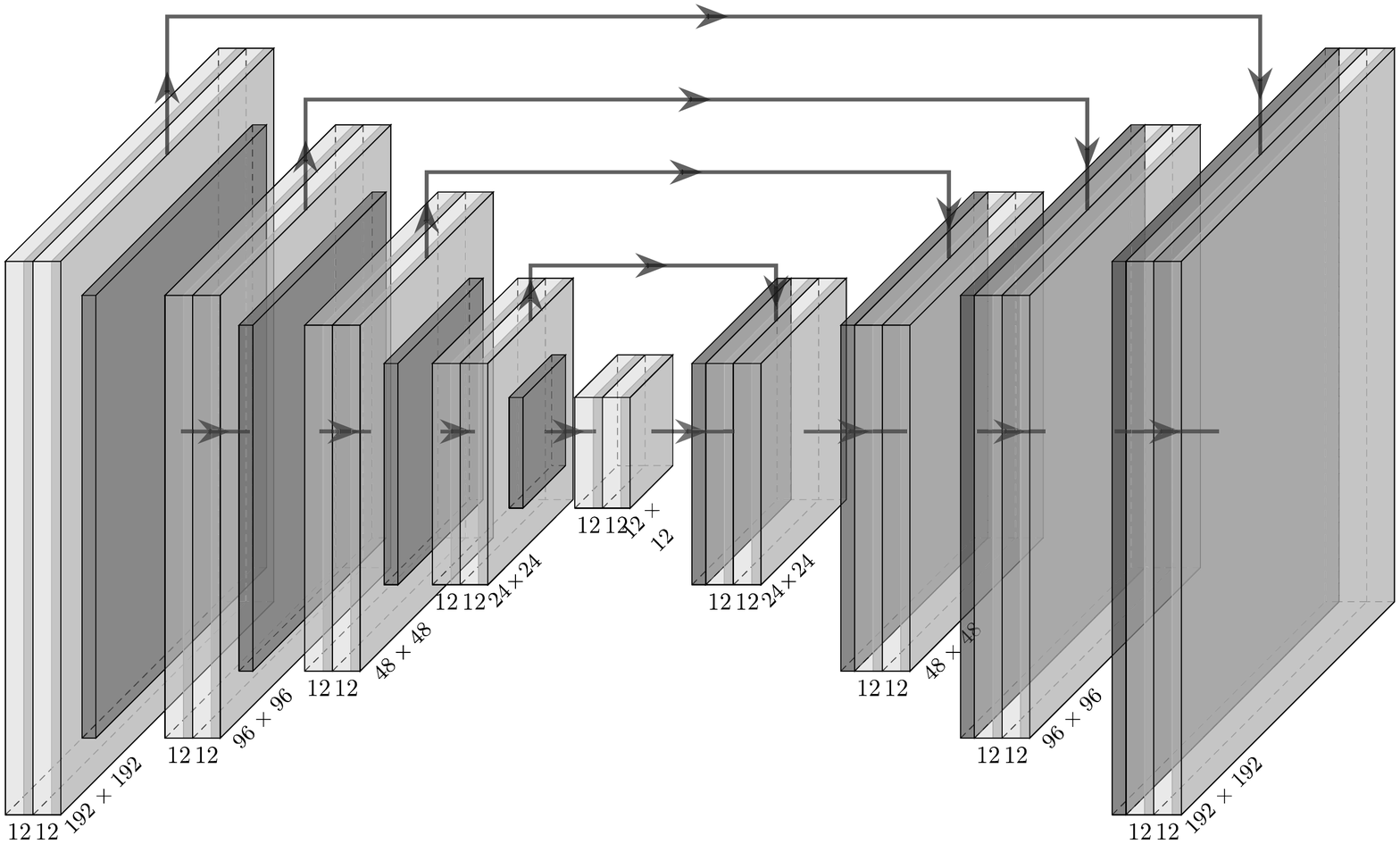}
	\caption{$\text{V-net}^{^{D,C}}_\Theta$ with parameters $\Theta$, depth $D=4$ and $C=12$ channels for images of resolution $N\times N=192 \times 192$.}
	\label{fig:V-net}
\end{figure}

For this purpose we propose a new network structure, similar to FlowNetSimple \cite{fischer2015}, general enough to produce global motion estimations but parameterized only by few parameters to prevent the network from overfitting. This U-net \cite{ronneberger2015} like structure with depth $D$ and a depth-independent number of channels $C$, here and in the following denoted by V-net, is visualized in Figure \ref{fig:V-net}. The depth is chosen according to $D = \lfloor \log_{2}(N/3)-1 \rfloor$ where $N$ is the discrete resolution of the $N\times N$ image. The activation is chosen as leaky-ReLU with negative slope coefficient $\alpha=0.1$.

To allow the network to refine an initial estimate of the deformation, the whole process of motion estimation is executed in an unrolled iteration scheme. The authors in \cite{ilg2016} received with this approach crisp results for optical-flow estimation. Our approach is summarized in Algorithm \ref{alg:motion_estimation}.
\begin{algorithm}
	\caption{Motion Estimation}
	\label{alg:motion_estimation}
	\begin{algorithmic}[1]
		\STATE{$\mathrm{Input}:  \, \, s_i, \,\, i=1,\ldots,N^\mathrm{exc}$} 
		\STATE{$\mathrm{Output}:  \, \, U_i^{\text{est}}:= U_i^{(N^{\mathrm{iter}}+1)}, \,\, i=1,\ldots,N^\mathrm{exc}$}
		\FOR{ $k=1,\ldots,N^\mathrm{iter}$}                    
		\FOR{ $i = 2, \ldots, N^\mathrm{exc}$}   
		\STATE {$\Bar{U}^{(k+1)}_i = U^{(k)}_i + \,\mathbin{\textbf{\text{V-net}}^{D,8}_{\Theta_i^{(k)}}}(U^{(k)}_i[s_1], \, s_i)$} 
		\ENDFOR
		\STATE {$U^{(k+1)} = \mathbin{\textbf{\text{V-net}}^{D-1,4N^\mathrm{exc}}_{\Theta_0^{(k)}}}(\Bar{U}^{(k+1)})$}
		\ENDFOR
	\end{algorithmic}
\end{algorithm}

Herein, the last $\text{V-net}_{\Theta_0}$ enables the network to include correlations between $U_i$ and hence time-consistency. The number of channels has to be increased for this task. A number of $C=4N^\mathrm{exc}$ channels were chosen. The iteration number was set to $N^\mathrm{iter}=4$. This network will in the following be referred to as the estimation network (est-net).

\subsection{Local Motion Correction}

As mentioned in \ref{subsec:motion_correction}, an analytic reconstruction will in general lead to additional artifacts stemming from inaccuracies in $U^\mathrm{est}$. For motion correction we therefore aim to use the projected gradient descent procedure to achieve admissible deformations. Instead of modelling the projection operators explicitly, we rely on Deep CNNs. 

All equations presented in \ref{subsec:motion_correction} are not limited to a specific resolution of the data. Since for radial acquisitions each excitation contains the information of the central Fourier coefficients, each excitation is represented at lower resolutions as well. Therefore taking the central Fourier coefficients only by cropping $y_i^c \subset \mathbb{C}^{N\times N}$ to $y_i^{c,\text{crop}=h} \subset \mathbb{C}^{N/2^h\times N/2^h}$ leads to a problem of resolution $N/2^h\times N/2^h$. This allows to 
compute the residuum and gradients at this lower resolution. Hence, this immediately suggests to use motion correction in a multilevel setup, starting at low resolutions and then refining the correction up to the full resolution. 

The procedure is presented in Algorithm \ref{alg:motion_correction}.
\begin{algorithm}
	\caption{Motion Correction}
	\label{alg:motion_correction}
	\begin{algorithmic}[1]
		\STATE{$\mathrm{Input}:  \, \, y_i, U_i^{\mathrm{est}}, \,\, i=1,\ldots,N^\mathrm{exc}$} 
		\STATE{$\mathrm{Output}:  \, \, U_i^\mathrm{cor}:= U_i^{(N^{\mathrm{iter}}+1,0)}, \,\, i=1,\ldots,N^\mathrm{exc}$}
		\STATE{$\mathrm{Initialize}: s^{(1,L)} = 0, \,\, U^{(1,L)}=\mathrm{Interpol}(U^{\mathrm{est}}, L) $}
		\FOR{ $h=L,L-1,\ldots,1,0$}
		\IF{$h<L$}
		\STATE{$ U^{(1,h)} = \mathrm{Extrapol}(U^{(N^\mathrm{iter}+1,h+1)}, h)$}            
		\STATE{$ s^{(1,h)} = \mathrm{Extrapol}(s^{(N^\mathrm{iter}+1,h+1)}, h)$}            
		\ENDIF
		\STATE{$ y^{\text{crop}=h} = \mathrm{Crop}(y, h)$}            
		
		\FOR{ $k=1,\ldots,N^\mathrm{iter}$}
		\STATE{$ s^{(k+1,h)} = \mathrm{CG}(y^{\text{crop}=h},U^{(k,h)},s^{(k,h)}, N^\mathrm{CG})$}            
		\FOR{ $i = 2, \ldots, N^\mathrm{exc}$}   
		
		\STATE {$U^{\Delta}_i = 
			$ \parbox[t]{0.7\linewidth}{%
				$\Bar{H}(U^{(k,h)})^{-1}\nabla_{U_i} J(U^{(k,h)}, s^{(k+1,h)}) + {\textbf{\text{V-net}}^{D-h,48}_{\Phi^{(k,h)}}}($ \\
				$\Bar{H}(U^{(k,h)})^{-1}\nabla_{U_i} J(U^{(k,h)}, s^{(k+1,h)}), $ \\ 
				$\Bar{H}(U^{(k,h)})^{-1}\nabla_{U_1} J(U^{(k,h)}, s^{(k+1,h)}), $ \\
				$U_i^{(k,h)}[s^{(k+1,h)}])$\;}} 
		\STATE {$\Bar{U}^{(k,h)}_i = U^{(k,h)}_i - U^{\Delta}_i$}
		
		\STATE {$U^{(k+1,h)}_i = $ \parbox[t]{0.7\linewidth}{%
				$\Bar{U}^{(k,h)}_i + \mathbin{\textbf{\text{V-net}}^{D-h-1,24}_{\Psi^{(k,h)}}}(\Bar{U}^{(k,h)}_i, U^{\Delta}_i, U_i^{(k,h)}[s^{(k+1,h)}])$\;}} 
		\ENDFOR
		\ENDFOR
		\ENDFOR
	\end{algorithmic}
\end{algorithm}

Herein $\mathrm{CG}(y,U,s,N^\mathrm{CG})$ is the CG-SENSE plus motion algorithm with data $y$, deformations $U$, initial value $s$ and iteration number $N^\mathrm{CG}$. We additionally use the positivity constraint of the solution and employ the Polak–Ribière version for nonlinear CG \cite{polak1969}. Note, that the parameters $\Phi^{(k,h)}$ and $\Psi^{(k,h)}$ are shared w.r.t the excitation number $i$. The extrapolation operator $ \mathrm{Extrapol}(s, h)$ extrapolates $s$ using a bilinear basis to the resolution $N/2^h\times N/2^h$. The interpolation operator $\mathrm{Interpol}(s, h)$ works in the opposite direction. $\nabla_{U_i} J(U^{(k,h)})$ is defined in equation \eqref{eq:gradients} and the approximate Hessian $\Bar{H}$ is given in equation \eqref{eq:final_Hessian}. The parameter $N^\mathrm{CG}$ can be adapted to the characteristics of the data. In our case we fix the parameters $N^\mathrm{iter}=2$ and $L=2$.

We split the projection of the gradients into two networks $\text{V-net}_{\Phi^{(k,h)}}$ and \\
$\text{V-net}_{\Psi^{(k,h)}}$. The reason is that local optimization techniques like the Adam optimizer always end up at local optima. Since the actual gradients provide rather sparse information a single network with full inputs of $U_i^{(k,h)}$, $s^{(k+1,h)}$ and the gradients would rather rely on $U_i^{(k,h)}$ and $s^{(k+1,h)}$ to correct inaccuracies leaving the gradients unused. This would lead to worse performance and even enhance overfitting. By splitting these tasks we enforce that all parameters of the first V-Net are dedicated to the gradients. In this sense it is ensured that the sparse nature of the gradients can not lead to a disregard of that information due to the training process.

Note that each instance of $\text{V-net}_{\Phi^{(k,h)}}$ and $\text{V-net}_{\Psi^{(k,h)}}$ receives arguments from one excitation only at a time. Therefore this network structure will work even if temporally uncorrelated deformations occur. This is particularly important if the temporal model for training is not rich enough to capture effects like heart beat introduced motions on top of breathing for in vivo experiments. As long as the spatial model captures these deformations, motion correction can compensate such high frequent effects.

Characteristic motion artifacts in $s^{(k)}$ like soft edges or streaking artifacts are actually necessary for intermediate reconstructions within the iterations of Algorithm \ref{alg:motion_correction}, such that gradients w.r.t. $U$ are meaningfull and the estimation on $U$ can be improved. Hence, a too perfect reconstruction rightaway would limit the power of the network. Therefore no denoiser is applied during the iterative process.

These insights suggest that, in general, motion correction and final reconstruction should be split for more reliable results.

In the following, we refer to this network as the correction network (cor-net).

\subsection{Experimental Setup}
The Network is aimed to be trained in a supervised learning fashion. Therefor training data are generated synthetically.

The synthetic deformation field is uncorrelated to the image itself. Therefore it is in general unequal to zero even in regions where no density gradient of the image is present. This way we can expect that in these regions, still the deformation field with highest probability is approximated. Therefore we can expect a visually similar solution known from sparse inpainting 
or the solution of Laplace equation with sparse boundary conditions \cite{weickert1996}.
This is particularly important for medical images where over large regions almost no density gradient is present as it appears in the liver or lung. That way we can hope to produce images that allow an accurate diagnosis in these regions by recovering marginal details.

The images are paired with artificially generated deformations. To prohibit specifically overfitting, we pair one image with several deformations. Due to this construction networks cannot directly relate an image to a deformation by the visual appearance but only indirectly by deducing misregistration or in a more subtle way.

The application of non-uniform fast Fourier transforms (NUFFT) seamlessly suits our formulations. Usually density compensation is used to compensate the heavy oversampling of the center of k-space at radial acquisitions. We therefore employ the density compensation proposed by Pipe et al. \cite{pipe1999} and use it excitation wise. By replacing $A_i\mathcal{F}$ with $D_i^{\frac{1}{2}}\mathcal{F}^\mathrm{NU}_i$, where $D_i$ is the density compensation and $\mathcal{F}^\mathrm{NU}_i$ is the NUFFT corresponding to time $t_i$, all equations are mutatis mutandis valid. 

The downside of NUFFT is the time-consuming computation. Since we only rely on gradients, it is their appearance that matters. Even though different forward operator models show differently pronounced artifacts within the reconstruction $s$, the gradients are very similar since the artifacts are equilibrated w.r.t. the respective forward operator. For training we therefore simulate and reconstruct using DFFT and evaluate the nearest k-space point on the regular grid. That way the operator $A_i$ is a diagonal matrix with ones and zeros on its diagonal corresponding to sampled and not sampled k-space points respectively. Therefore the training is very fast and allows to train the correction network on larger data sets. In that sense we can train the network using the DFFT and transfer the weights to other forward operator models.

\subsection{Network Training}
For the network training we fix the resolution to $N=192$. To cover the k-space appropriately using radial acquisition requires $192$ uniformly distributed spokes in total. This is already a factor of $\pi/2$ below Nyquist requiring the SENSE information to allow recovering the true image in a static sense. Hence for $N^\mathrm{exc}=16$ each undersampled CG-SENSE reconstruction consists of $12$ uniformly distributed spokes. The Nyquist criterion is therefore in this case violated by a factor of $8\pi$. 

During training, dropout \cite{srivastava2014} layers are used for better noise-resistance.

\subsubsection{Motion Estimation}

The estimation network was trained with Adam optimizer and MSE loss for approximately 300 epochs with 4800 training pairs and a batch-size of at maximum $16$ suited to the memory limits. Additionally batch normalization was used. The estimation network relies on the appearance of the undersampled CG-SENSE reconstructions and the network architecture evolves with $N^\mathrm{exc}$. In that sense the estimation network needs to be trained for each required $N^\mathrm{exc}$. Interchanging the forward operator model to NUFFT for evaluation showed good performance. Therefore the network was not separately adapted to NUFFT reconstructions knowing that this could further improve the quality of the motion estimates. The estimation network has 4.302 million trainable parameters for $N^\mathrm{exc}=16$.

\subsubsection{Motion Correction}
Motion correction was trained on a set of motion estimates generated by the estimation network. This yields the inputs $(y,U^\mathrm{est})$. 

The correction network was trained with Adam optimizer and loss function 
\begin{align}
	\mathcal{L}^\mathrm{cor} = & \|(U-U^\mathrm{ref})_{|s^\mathrm{ref}>0} + 0.1(U-U^\mathrm{ref})_{|s^\mathrm{ref}=0} \|_2^2 + \nonumber \\
	&\beta_1 \|R(U,s)\|_2^2 + \beta_2(\|\partial_{xx}U\|_2^2 + \|\partial_{yy}U\|_2^2) \nonumber 
\end{align}
on a data set of 4800 training pairs. Herein $R$ denotes the residuum defined in equation \eqref{eq:residuum}. The hyperparameters $\beta_1$ and $\beta_2$ were chosen by hand such that the terms are approximately weighted equally. It was found that an end-to-end training showed instabilities during training. Therefore a cascade training procedure \cite{marquez2018} was employed. Within each cascade the network fragments $\text{V-net}_{\Phi^{(k,h)}}$ and $\text{V-net}_{\Psi^{(k,h)}}$ are trained together. This enables fast gradient computations and memory efficiency since no gradient flow across CG or the computation of $\nabla_{U_i}J(U^{(k,h)})$ is needed. This allows to train the correction network with a maximum batch size of 8 suited to the memory limits. The correction network has 3.972 million trainable parameters in total.

Since the correction network relies on gradients, no batch normalization can be used because this would destroy essential properties like the sign and the magnitude of the gradient. Without  batch normalization, the networks do not generalize well if data of severely different magnitude are introduced. In that sense some linear scaling needs to be applied such that the data sample matches the data distribution of the training data set.

\section{Results}

For our tests we synthetically construct two test cases. 

\begin{figure}
	\centering
	\begin{subfigure}[b]{0.25\textwidth}
		\centering
		\begin{tikzpicture}[spy using outlines={white,magnification=3,size=1.5cm, connect spies}]
			\node {\includegraphics[width=1.0\textwidth]{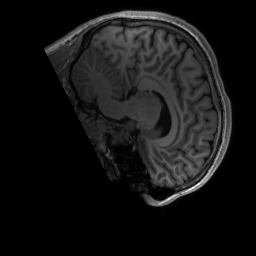}};
			\spy on (1.25,0.0) in node [left] at (-0.25,-1.0); 
		\end{tikzpicture}
		\caption*{ground truth $s^\mathrm{ref}$}
	\end{subfigure}
	\hspace{0.15\textwidth}
	\begin{subfigure}[b]{0.25\textwidth}
		\begin{tikzpicture}[spy using outlines={white,magnification=3,size=1.5cm, connect spies}]
			\node {\pgfimage[width=1.0\textwidth]{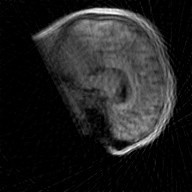}};
			\spy on (1.25,0.0) in node [left] at (-0.25,-1.0);
		\end{tikzpicture}

		\caption*{static reconstruction}
	\end{subfigure}
	\\
	\begin{subfigure}[b]{0.25\textwidth}
		\begin{tikzpicture}[spy using outlines={white,magnification=3,size=1.5cm, connect spies}]
			\node {\pgfimage[height=1.0\textwidth]{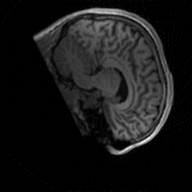}};
			\spy on (1.25,0.0) in node [left] at (-0.25,-1.0);
		\end{tikzpicture}
		
		\caption*{$\text{} \qquad \, \, \text{CG reconstruction } s^\mathrm{est}$ \\ $\text{} \qquad \qquad \quad \text{with } U^{\mathrm{est}}$}
	\end{subfigure}
	\hspace{0.15\textwidth}
	\begin{subfigure}[b]{0.25\textwidth}
		\begin{tikzpicture}[spy using outlines={white,magnification=3,size=1.5cm, connect spies}]
			\node {\pgfimage[height=1.0\textwidth]{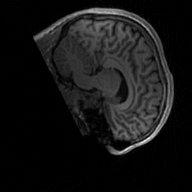}};
			\spy on (1.25,0.0) in node [left] at (-0.25,-1.0);
		\end{tikzpicture}
		\caption*{$\text{} \qquad \, \, \text{CG reconstruction } s^\mathrm{cor}$ \\ $\text{} \qquad \qquad \quad \text{with } U^{\mathrm{cor}}$}
	\end{subfigure}
	\\
	\begin{subfigure}[b]{0.25\textwidth}
	
		\begin{tikzpicture}[spy using outlines={white,magnification=3,size=1.5cm, connect spies}]
			\node {\pgfimage[height=1.0\textwidth]{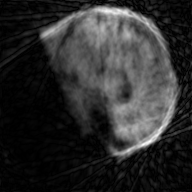}};
			\spy on (1.25,0.0) in node [left] at (-0.25,-1.0);
		\end{tikzpicture}
		\caption*{$\text{} \quad \, \, \, \, \text{undersampled CG-SENSE}$\\  $\text{} \qquad \quad \, \, \text{reconstruction } s_{13}$}
	\end{subfigure}
	\hspace{0.15\textwidth}
	\begin{subfigure}[b]{0.25\textwidth}
	
		\begin{tikzpicture}[spy using outlines={white,magnification=3,size=1.5cm, connect spies}]
			\node {\pgfimage[height=1.0\textwidth]{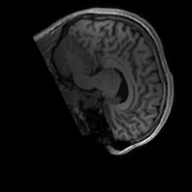}};
			\spy on (1.25,0.0) in node [left] at (-0.25,-1.0);
		\end{tikzpicture}
		\caption*{$\text{} \quad \text{final reconstruction } s^\mathrm{cor} \text{ after} $ \\ $\text{} \qquad \quad \text{denoising with TV }$}
	\end{subfigure}
	\\
	\begin{subfigure}[b]{0.315\textwidth}
		\begin{tikzpicture}[spy using outlines={white,magnification=3,size=1.5cm, connect spies}]
			\node {\pgfimage[width=0.8\textwidth]{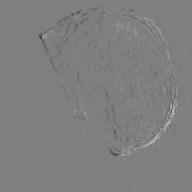}};
			\hspace{2.7cm}
			\node 		{\pgfimage[width=0.18\textwidth,height=0.8\textwidth]{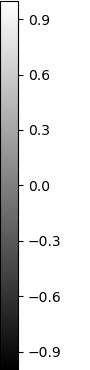}};
		\end{tikzpicture}
		\caption*{$(\Bar{H}_{13}^y)^{-1}\dd J(U)/\dd p_{13}^y$ at \\ $U^{\mathrm{est}}$, $s^\mathrm{est}$}
	\end{subfigure}
	\hspace{0.1\textwidth}
	\begin{subfigure}[b]{0.25\textwidth}
		\begin{tikzpicture}[spy using outlines={white,magnification=3,size=1.5cm, connect spies}]
			\node {\pgfimage[width=1.0\textwidth]{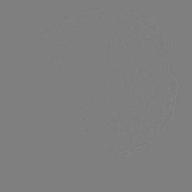}};
		\end{tikzpicture}
		\caption*{$\text{} \qquad (\Bar{H}_{13}^y)^{-1}\dd J(U)/\dd p_{13}^y$ at \\ $\text{} \qquad \qquad \quad U^{\mathrm{cor}}$, $s^\mathrm{cor}$}
	\end{subfigure}

\caption{Reconstructions using NUFFT from noise-free synthetically motion corrupted Brain Yale data set with $N^\mathrm{exc}=16$, $N^\mathrm{coils}=4$. Weights are unchanged from training with DFFT. }
\label{fig:reconstructions_synthetic_Brainweb}
\end{figure}

The first test case is the Yale High-Resolution Controls Dataset \cite{finn2015} of $256\times256$ ground truth MPRAGE Brain MRIs coupled with affine deformations. These affine deformations are chosen with rotations of up to $\pm 10^\circ $ and shifts of up to $\pm 3\%$ of the length of $\Omega^\mathrm{FOV}$. The temporal correlation is suited to continuous time GRE acquisitions with a repetition time of $\approx 0.002s-0.01s$ where at each repetition one spoke is acquired. In addition the NUFFT is used as the forward operator. The results are shown in Figure \ref{fig:reconstructions_synthetic_Brainweb}.

The results on Brain Yale data are produced without explicit adaption to the NUFFT. The proposed method eliminates virtually all deformation artifacts and produces reliable reconstructions and deformation estimates. In our simulations we did not distinguish if during motion the brain partially left the FOV. 

The brain itself as well as the exterior skull are reconstructed particularly well. Already the estimation network alone is for such simple deformations capable of producing almost motion artifact free reconstructions.

Altogether, the learned networks generalize well to unseen data and even unseen forward operators. It has to be emphasized, that if trained to NUFFT, the results are even superior. In that sense we expect that if trained to synthetic NUFFT data, the networks will generalize better to in vivo MRI data. Quantitative results are presented in Table \ref{tab:quality_of_reconstructions_brainweb}. The quality of the final reconstruction is assessed using different error metrics, namely SSIM (structure similarity index), PSNR (peak signal to noise ratio) and MSE (mean squared error). Additionally the squared norm of the residuum $\mathrm{Res} = \|R\|_2^2$ is shown. We provide results on noise-free and noisy data. In the latter case we also state the static measure of incompatibility of the noise which is the residual value that remains even for an optimal reconstruction in the static case.

\begin{figure}
\centering
\hspace{-1.5cm}
\begin{subfigure}[b]{0.25\textwidth}
\centering
\begin{tikzpicture}[spy using outlines={white,magnification=3,size=1.5cm, connect spies}]
\node {\pgfimage[height=1.0\textwidth]{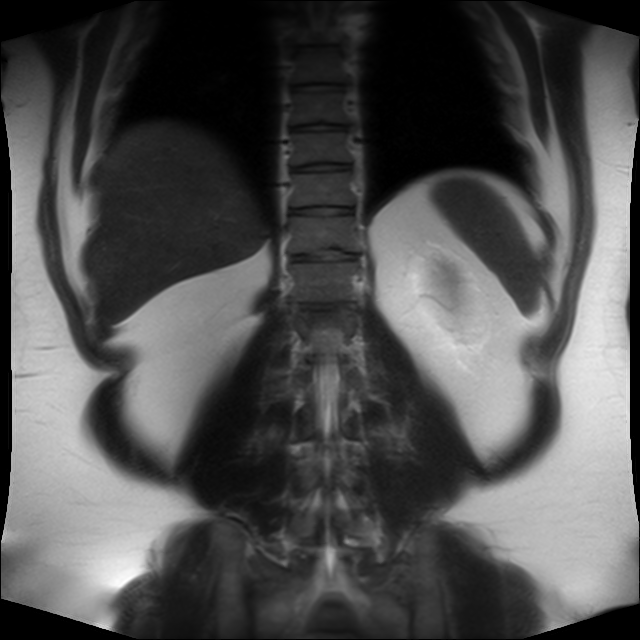}};
\spy on (0.15,0.2) in node [left] at (3.25,-1.0);
\end{tikzpicture}
\caption*{ground truth $s^\mathrm{ref}$}
\end{subfigure}
\hspace{0.1cm}
\begin{subfigure}[b]{0.25\textwidth}
\begin{tikzpicture}[spy using outlines={white,magnification=3,size=1.5cm, connect spies}]
\node {\pgfimage[height=1.0\textwidth]{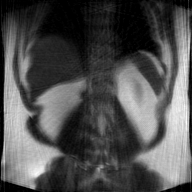}};
\spy on (0.15,0.2) in node [left] at (-1.75,1.0);
\end{tikzpicture}
\caption*{$\qquad \qquad \qquad \text{static reconstruction}$}
\end{subfigure}\hfill

\hspace{-1.5cm}
\begin{subfigure}[b]{0.25\textwidth}
\begin{tikzpicture}[spy using outlines={white,magnification=3,size=1.5cm, connect spies}]
\node {\pgfimage[height=1.0\textwidth]{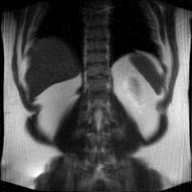}};
\spy on (0.15,0.2) in node [left] at (3.25,-1.0);
\end{tikzpicture}
\caption*{$s^\mathrm{est}$}
\end{subfigure}
\hspace{0.1cm}
\begin{subfigure}[b]{0.25\textwidth}
\centering
\begin{tikzpicture}[spy using outlines={white,magnification=3,size=1.5cm, connect spies}]
\node {\pgfimage[height=1.0\textwidth]{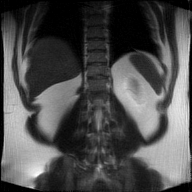}};
\spy on (0.15,0.2) in node [left] at (-1.75,1.0);
\end{tikzpicture}
\caption*{$\qquad \qquad \qquad \qquad s^\mathrm{cor}$}
\end{subfigure}\hfill

\hspace{-1.5cm}
\begin{subfigure}[b]{0.25\textwidth}
\centering
\begin{tikzpicture}[spy using outlines={white,magnification=3,size=1.5cm, connect spies}]
\node {\pgfimage[height=1.0\textwidth]{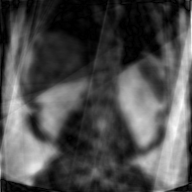}};
\spy on (0.15,0.2) in node [left] at (3.25,-1.0);
\end{tikzpicture}
\caption*{CG-SENSE reconstruction \\ $s_{13}$}
\end{subfigure}
\hspace{0.1cm}
\begin{subfigure}[b]{0.25\textwidth}
\centering
\begin{tikzpicture}[spy using outlines={white,magnification=3,size=1.5cm, connect spies}]
\node {\pgfimage[height=1.0\textwidth]{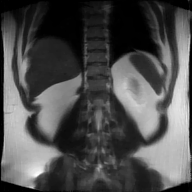}};
\spy on (0.15,0.2) in node [left] at (-1.75,1.0);
\end{tikzpicture}
\caption*{$\qquad \qquad \quad \,\,\, \text{final reconstruction } s^\mathrm{cor}$ \\ $ \text{} \qquad \qquad \quad \,\,\, \text{after denoising with TV}$}
\end{subfigure}\hfill

\caption{Reconstructions of the image from synthetic Abdominal test data set with $N^\mathrm{exc}=16$, $N^\mathrm{coils}=4$, high resolution $640\times640$ NUFFT model and time-continuous data acquisition. All reconstructions have a lower resolution of $192\times192$. Weights are unchanged from training with DFFT. }
\label{fig:reconstructions_synthetic_Abdominal_s}
\end{figure}

\begin{figure}
\centering
\begin{subfigure}[b]{0.25\textwidth}
\centering
\includegraphics[width=\textwidth]{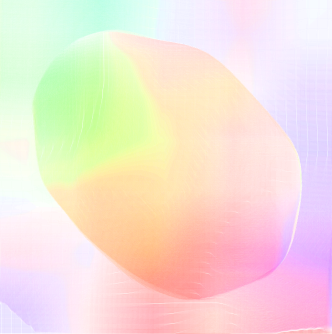}
\caption*{ground truth $U^{\mathrm{ref}}_{13}$}
\end{subfigure}
\hspace{0.15\textwidth}
\begin{subfigure}[b]{0.25\textwidth}
\centering
\includegraphics[width=\textwidth]{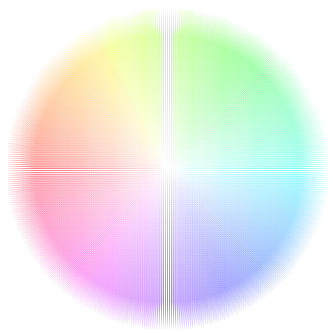}
\caption*{$\text{color map for visualization of } |U|$ \\ $\text{} \qquad \qquad \,\,\,\, \text{and } \arg U$}
\end{subfigure}
\\
\begin{subfigure}[b]{0.25\textwidth}
\centering
\includegraphics[width=\textwidth]{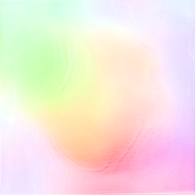}
\caption*{$U_{13}^\mathrm{est}$}
\end{subfigure}
\hspace{0.15\textwidth}
\begin{subfigure}[b]{0.25\textwidth}
\centering
\includegraphics[width=\textwidth]{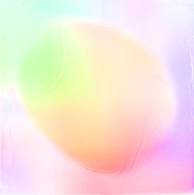}
\caption*{$U_{13}^\mathrm{cor}$}
\end{subfigure}
\\
\begin{subfigure}[b]{0.315\textwidth}
\begin{tikzpicture}
\node {\pgfimage[width=0.8\textwidth]{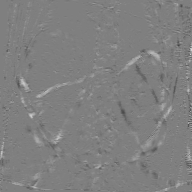}};
\hspace{2.7cm}
\node {\pgfimage[width=0.18\textwidth,height=0.8\textwidth]{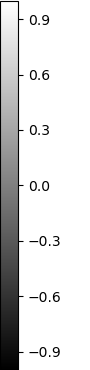}};
\end{tikzpicture}
\caption*{$(\Bar{H}_{13}^x)^{-1}\dd J(U)/\dd p_{13}^x$ at \\ $U^{\mathrm{est}}$, $s^\mathrm{est}$}
\end{subfigure}
\hspace{0.1\textwidth}
\begin{subfigure}[b]{0.25\textwidth}
\centering
\includegraphics[width=\textwidth]{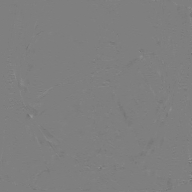}
\caption*{$\text{} \qquad (\Bar{H}_{13}^x)^{-1}\dd J(U)/\dd p_{13}^x$ at \\ $\text{} \qquad \qquad \quad  \text{} U^{\mathrm{cor}}$, $s^\mathrm{cor}$}
\end{subfigure}
\caption{Reconstructions of the deformation fields from synthetic Abdominal test data set with $N^\mathrm{exc}=16$, $N^\mathrm{coils}=4$, high resolution $640\times640$ NUFFT model and time-continuous data acquisition. All reconstructions have a lower resolution of $192\times192$. Weights are unchanged from training with DFFT. }
\label{fig:reconstructions_synthetic_Abdominal_U}
\end{figure}

The second test case comprises high resolution abdominal MRIs of 38 subjects provided by the Uniklinikum Würzburg. We simulate a continuous time data acquisition
corrupted with synthetic free form deformations as proposed in Subsection \ref{subsec:subsection_deformation_model} and the NUFFT forward operator. Therefore the k-space data are sampled with a bandwidth conform with the training resolution. The results are shown in Figure \ref{fig:reconstructions_synthetic_Abdominal_s} and \ref{fig:reconstructions_synthetic_Abdominal_U}.

The results surpass the quality of adversarial techniques. While adversarial performance significantly abates at strong deformation artifacts \cite{usman2020}, our results can be expected to be basically independent of the magnitude of the deformation due to the use of the estimation network. Quantitative results are presented in Table \ref{tab:quality_of_reconstructions_abdominal}.

\begin{figure}
\centering
\begin{subfigure}[b]{0.25\textwidth}
\begin{tikzpicture}[spy using outlines={white,magnification=3,size=1.5cm, connect spies}]
\node {\pgfimage[height=1.0\textwidth]{figures/feinl7.png}};
\spy on (1.25,0.0) in node [left] at (-0.25,-1.0);
\end{tikzpicture}
\caption*{ours}
\end{subfigure}
\hspace{0.05cm}
\begin{subfigure}[b]{0.25\textwidth}
\centering

\begin{tikzpicture}[spy using outlines={white,magnification=3,size=1.5cm, connect spies}]
\node {\pgfimage[height=1.0\textwidth]{figures/feinl16.png}};
\spy on (0.15,0.2) in node [left] at (-1.75,1.0);
\end{tikzpicture}
\caption*{$\qquad \qquad \qquad \qquad \quad \mathrm{ours}$}
\end{subfigure}\hfill
\\
\begin{subfigure}[b]{0.25\textwidth}
\begin{tikzpicture}[spy using outlines={white,magnification=3,size=1.5cm, connect spies}]
\node {\pgfimage[height=1.0\textwidth]{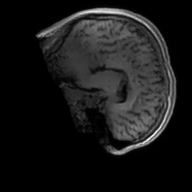}};
\spy on (1.25,0.0) in node [left] at (-0.25,-1.0);
\end{tikzpicture}
\caption*{GAN}
\end{subfigure}
\hspace{0.05cm}
\begin{subfigure}[b]{0.25\textwidth}
\centering

\begin{tikzpicture}[spy using outlines={white,magnification=3,size=1.5cm, connect spies}]
\node {\pgfimage[height=1.0\textwidth]{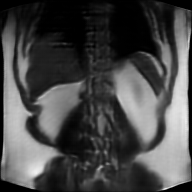}};
\spy on (0.15,0.2) in node [left] at (-1.75,1.0);
\end{tikzpicture}
\caption*{$\qquad \qquad \qquad \qquad \quad \mathrm{GAN}$}
\end{subfigure}\hfill
\\
\begin{subfigure}[b]{0.25\textwidth}
\begin{tikzpicture}[spy using outlines={white,magnification=3,size=1.5cm, connect spies}]
\node {\pgfimage[height=1.0\textwidth]{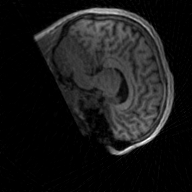}};
\spy on (1.25,0.0) in node [left] at (-0.25,-1.0);
\end{tikzpicture}
\caption*{rigid}
\end{subfigure}
\hspace{0.05cm}
\begin{subfigure}[b]{0.25\textwidth}
\centering
\begin{tikzpicture}[spy using outlines={white,magnification=3,size=1.5cm, connect spies}]
\node {\pgfimage[height=1.0\textwidth]{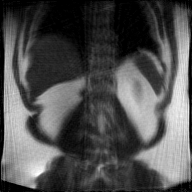}};
\spy on (0.15,0.2) in node [left] at (-1.75,1.0);
\end{tikzpicture}
\caption*{$\qquad \qquad \qquad \qquad \quad \mathrm{rigid}$}
\end{subfigure}\hfill
\caption{Visual comparison of reconstructions of our method with reconstructions from GAN (Usman) and rigid motion estimation (Cordero). The first column compares the results on the synthetic Brain Yale data set, the second column provides results on synthetic Abdominal data.}
\label{fig:reconstructions_GAN_and_cordero}
\end{figure}

A visual comparison is shown in Figure \ref{fig:reconstructions_GAN_and_cordero}. Note, that even though the GAN reconstruction produces high error norms, the visual appearance seems much more pleasing than a static reconstruction. A closer look at the reconstructed brain reveals that rather some artificial brain alike image is generated than actual deformation is removed.

Even though the data are generated by a continuous acquisition, the deformation estimates at discrete times are used for final reconstruction. This might introduce slight temporal blurring which can be fixed by increasing $N^\mathrm{exc}$ of the motion estimation procedure, or just the extrapolation to continuous deformations over time. Since this requires a gridding operation for each spoke, the time for final reconstruction drastically increases. However, in our tests, such temporal blurring appeared to be negligible.

In all cases we use $N^\mathrm{coils}=4$ coils constructed by $S_c(r)=\exp(-((r-r_c)^2)/d_c)$ with $r_c$ and $d_c$ chosen randomly in appropriate intervals.

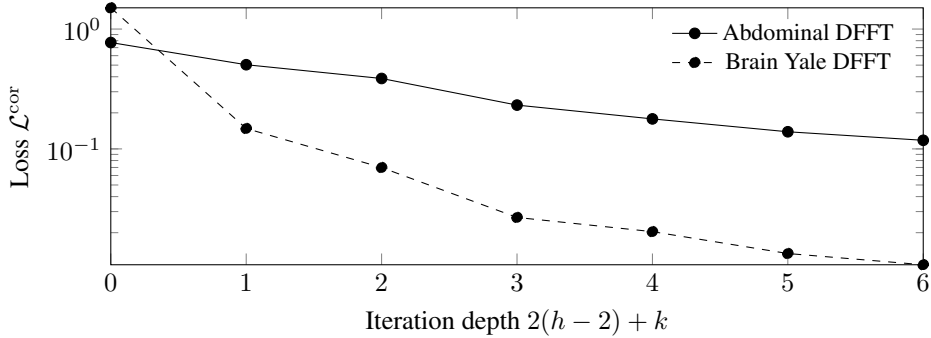
\begin{figure}
\centering
\begin{tikzpicture}

\begin{semilogyaxis}[
xmin = 0, xmax = 6,
ymin = 0.0108, ymax = 1.504,height=5cm,width = 0.75\textwidth,
xlabel=Iteration depth $2(h-2)+k$,
ylabel=Loss $\mathcal{L}^\mathrm{cor}$,
legend style={nodes={scale=0.9},draw=none,fill=none}]
\addplot[black, mark=*] table[x=iter, y=adaptAbdom3_test] {B.dat};
\addplot[black, mark=*, dashed] table[x=iter, y=adaptBrain_test] {B.dat};
\legend{Abdominal DFFT, Brain Yale DFFT}
\end{semilogyaxis}

\end{tikzpicture}
\caption{Test Error Loss performance of Correction Network over iteration depth with $N^\mathrm{exc}=16$. In analogy to Algorithm \ref{alg:motion_correction} we use $h$ and $k$ as indices to describe the cascade. At iteration number zero the loss of the result from estimation network is shown. }
\label{fig:Cor_Net_performance}
\end{figure}

In Figure \ref{fig:Cor_Net_performance} the convergence of the loss of the correction network is shown over iteration depth. We use the synthetic Brain Yale test data set and the synthetic Abdominal test data set for visualization of the error.

The computation time of the whole procedure consists of three parts, namely the computation of the undersampled CG reconstructions $s_i$, motion estimation and motion correction. The computation of $\{s_i\}$ scales linearly with the number of excitations, and of course the computation time of the forward operator. Motion estimation asymptotically scales quadratically with $N^\mathrm{exc}$ but the computation times are almost negligible in context to the whole procedure. Motion correction scales linearly with the $N^\mathrm{exc}$ and the computation time of the forward operator. In Table \ref{tab:computation_times} explicit computation times are shown for a workstation with Intel i9-10900X, 128GB Ram and NVIDIA RTX 3090. In our implementation we used an adapted tensorflow version of the NUFFT of Muckley et al. \cite{muckley2020} which employs the fact that trajectories do not change during the iterative procedure. Our GPU implementation parallelizes particularly well over the number of coils leading to an increase of only 10\% computation time when using $N^\mathrm{coils}=16$ coils.
The use of a tensorflow implementation of trajectory optimized NUFFT could further decrease NUFFT computation times significantly.

\begin{table}
\centering
\begin{tabular}{ l||ll|ll|ll|ll|ll } 
\hline
\hline

Method & \multicolumn{2}{c}{$\{s_i\}$} & \multicolumn{2}{c}{est net} & \multicolumn{2}{c}{cor net} & \multicolumn{2}{c}{Cordero} & \multicolumn{2}{c}{Usman}\\
$N^\mathrm{exc}$ & 8 & 16 & 8 & 16 & 8 & 16 & 8 & 16 & 8 & 16 \\
\hline

\hline

DFFT         & 0.166 & 0.234 & 0.27 & 0.48 & 1.54 & 2.91 & 93.8 & 184 & 0.043 & 0.043 \\
NUFFT        & 2.66 & 5.24 & 0.27 & 0.48 & 6.9 & 13.5 & 179 & 364 & 0.043 & 0.043 \\

\hline
\hline
\end{tabular}
\caption{Computation time in seconds for one data sample and $N^\mathrm{coils}=4$. The computation time of Usman does not include the computation of the static reconstruction (approx. 0.4s for NUFFT and 0.03s for DFFT). }
\label{tab:computation_times}
\end{table}

\begin{table}
\centering
\begin{tabular}{ l|cccc } 
\hline

\hline

Method & Res & PSNR & SSIM & MSE \\
\hline
\multicolumn{5}{c}{noise level = 0\%} \\
\hline

static & 140.4 & 30.60 & 0.9001 & 1.152\\

rigid (Cordero)& 26.49 & 38.69 & 0.9708 & 0.2716 \\
GAN (Usman) & - & 29.37 & 0.8990 & 1.3873 \\
est-net & 9.943 & 40.94 & 0.9886 & 0.1066	\\
est- \& cor-net & \textbf{1.911} & 41.18 & 0.9900 & 0.1009 \\
est- \& cor-net \& TV & 2.219 & \textbf{42.01} & \textbf{0.9924} & \textbf{0.0883}\\
\hline

\multicolumn{5}{c}{noise level = 5\%} \\
\hline

static & 165.6 & 30.45 & 0.8958 & 1.170\\
rigid (Cordero)& 56.28 & 37.34 & 0.9592 & 0.3571 \\ 
GAN (Usman) & - & 29.36 & 0.8990 & 1.3895 \\
est-net & 35.64 & 40.27 & 0.9862 & 0.1183  	\\
est- \& cor-net & \textbf{27.18} & 40.25 & 0.9862 & 0.1173 \\
est- \& cor-net \& TV & 27.64 & \textbf{41.39} & \textbf{0.9910} & \textbf{0.0968} \\
\hline

\hline
\end{tabular}
\caption{Quality of reconstruction $s$ for synthetic Brain Yale data set with $N^\mathrm{exc}=16$ and NUFFT. The 5\% additive normal distributed measurement noise leads to a static measure of incompatibility of 19.0.}         
\label{tab:quality_of_reconstructions_brainweb}
\end{table}

\begin{table}
\centering
\begin{tabular}{ l|cccc } 
\hline

\hline

Method & Res & PSNR & SSIM & MSE \\
\hline
\multicolumn{5}{c}{noise level = 0\%} \\
\hline

static & 292.9 & 28.07 & 0.866 & 1.948\\
rigid (Cordero)& 134.83 & 30.74 & 0.907 & 1.127 \\
GAN (Usman)& - & 27.57 & 0.859 & 2.077\\
est-net & 20.65 & 36.24 & 0.969 & 0.296	\\
est- \& cor-net& \textbf{2.687} & 37.38 & 0.976 & 0.233\\
est- \& cor-net \& TV & 3.111 & \textbf{37.69} & \textbf{0.979} & \textbf{0.221}\\
\hline

\multicolumn{5}{c}{noise level = 5\%} \\
\hline

static & 422.3 & 27.55 & 0.833 & 2.140\\
rigid (Cordero)& 273.8 & 29.69 & 0.867 & 1.423 \\ 
GAN (Usman)& - & 27.34 & 0.850 & 2.182 \\
est-net & 154.5 & 34.74 & 0.949 & 0.4075 \\
est- \& cor-net & \textbf{130.9} & 34.45 & 0.940 & 0.4358 \\
est- \& cor-net \& TV & 132.2 & \textbf{35.58} & \textbf{0.957} & \textbf{0.3477} \\
\hline

\hline
\end{tabular}
\caption{Quality of reconstruction $s$ for synthetic Abdominal data set with $N^\mathrm{exc}=16$ and NUFFT. The 5\% additive normal distributed measurement noise leads to a static measure of incompatibility of 121.1.}
\label{tab:quality_of_reconstructions_abdominal}
\end{table}

\section{Conclusion}

The proposed procedure produces high quality reconstructions for motion affected multishot MRI. The results are superior to the state-of-the-art adversarial approaches of Usman et al. \cite{usman2020} and Armanious et al. \cite{armanious2018b} regarding image quality. 
In contrast to adversarial approaches, the results are additionally well-grounded by low residuum and the explicit calculation of deformation fields for each excitation. 

The computation time is due to the rather complex network structure and the need for CG iterations relatively long, compared to adversarial approaches. However, with an overall computation time of approximately 20 seconds (with a trajectory optimized NUFFT much faster computation times are achievable), a clinical application is very well feasible. In contrast, purely analytic methods \cite{burger2018} take significantly longer while the results are of similar quality than ours \cite{dirks2015}.

The precise calculation of sensitivity maps is essential for the performance of the proposed network. 
In analogy to Arvinte et al. \cite{arvinte2021} a joint iterative adaption of the maps could be combined with our network allowing even excitation dependent sensitivity map estimates. 

Since the network $\text{V-Net}_{\Omega^{(k,h)}}$ basically performs a sparse inpainting of the gradients, the network structure could be further inspired by state-of-the-art network architectures of inpainting \cite{lizuka2017} using e.g. dilated convolutions for better extrapolation.

\bibliographystyle{unsrt}  
\bibliography{Literatur}

\end{document}